\documentclass[conference]{IEEEtran}
\usepackage{cite}
\usepackage{amsmath,amssymb,amsfonts}
\usepackage{algorithmic}
\usepackage{graphicx}
\usepackage{textcomp}
\usepackage{xcolor}
\usepackage{booktabs}
\usepackage{cleveref}
\usepackage{tabularx}
\usepackage{makecell}
\usepackage{multirow}
\usepackage{multicol}
\def\BibTeX{{\rm B\kern-.05em{\sc i\kern-.025em b}\kern-.08em
    T\kern-.1667em\lower.7ex\hbox{E}\kern-.125emX}}
\begin{document}

\title{Temporal Object Captioning for Street Scene Videos from LiDAR Tracks\\
{\footnotesize \textsuperscript{}}
\thanks{Identify applicable funding agency here. If none, delete this.}
}

\author{\IEEEauthorblockN{1\textsuperscript{st} Vignesh Gopinathan}
\IEEEauthorblockA{\textit{Aptiv, Germany} \\
\textit{University of Wuppertal, Germany}}
\and
\IEEEauthorblockN{2\textsuperscript{nd} Urs Zimmermann}
\IEEEauthorblockA{\textit{Aptiv, Germany}}
\and
\IEEEauthorblockN{3\textsuperscript{rd} Michael Arnold}
\IEEEauthorblockA{\textit{Aptiv, Germany}}
\and
\IEEEauthorblockN{4\textsuperscript{th} Matthias Rottmann}
\IEEEauthorblockA{\textit{University of Wuppertal, Germany}}
}

\maketitle

\begin{abstract}
Video captioning models have seen notable advancements in recent years, especially with regard to their ability to capture temporal information. While many research efforts have focused on architectural advancements, such as temporal attention mechanisms, there remains a notable gap in understanding how models capture and utilize temporal semantics for effective temporal feature extraction, especially in the context of Advanced Driver Assistance Systems. 
We propose an automated LiDAR-based captioning procedure that focuses on the temporal dynamics of traffic participants. Our approach uses a rule-based system to extract essential details such as lane position and relative motion from object tracks, followed by a template-based caption generation.
Our findings show that training SwinBERT, a video captioning model, using only front camera images and supervised with our template-based captions, specifically designed to encapsulate fine-grained temporal behavior, leads to improved temporal understanding consistently across three datasets. In conclusion, our results clearly demonstrate that integrating LiDAR-based caption supervision significantly enhances temporal understanding, effectively addressing and reducing the inherent visual/static biases prevalent in current state-of-the-art model architectures.
\end{abstract}

\section{Introduction}
Video captioning has gained significant attention due to its ability to generate natural language descriptions from videos \cite{kuo2023mammutsimplearchitecturejoint,he2023vlabenhancingvideolanguage,chen2023valorvisionaudiolanguageomniperceptionpretraining}. These models typically leverage convolutional neural networks (CNNs) for visual feature extraction \cite{wang2019selfsupervisedspatiotemporalrepresentationlearning,caron2019unsupervisedpretrainingimagefeatures,jaworek2019melanoma,xie2018pre} and recurrent neural networks (RNNs) or transformers for caption prediction \cite{jin2019low,nakamura2021sensor,ullah2021boosting}. Beyond general video understanding, the capabilities of video captioning are gaining relevance in safety-critical domains such as Advanced Driver Assistance Systems (ADAS), where capturing dynamic scene context is essential. ADAS requires a comprehensive understanding of the surrounding environment to ensure safe and efficient navigation. Traditional computer vision tasks such as object detection, semantic segmentation, and motion prediction have been extensively utilized to understand visual scenes \cite{kirillov2019panoptic,divvala2009empirical,carion2020end}. However, video captioning introduces an additional layer of context by summarizing the complex and dynamic interactions between objects, actions, and scenes in a more holistic manner.
\begin{figure}[t]
    \centering
    \includegraphics[width=0.35\textwidth]{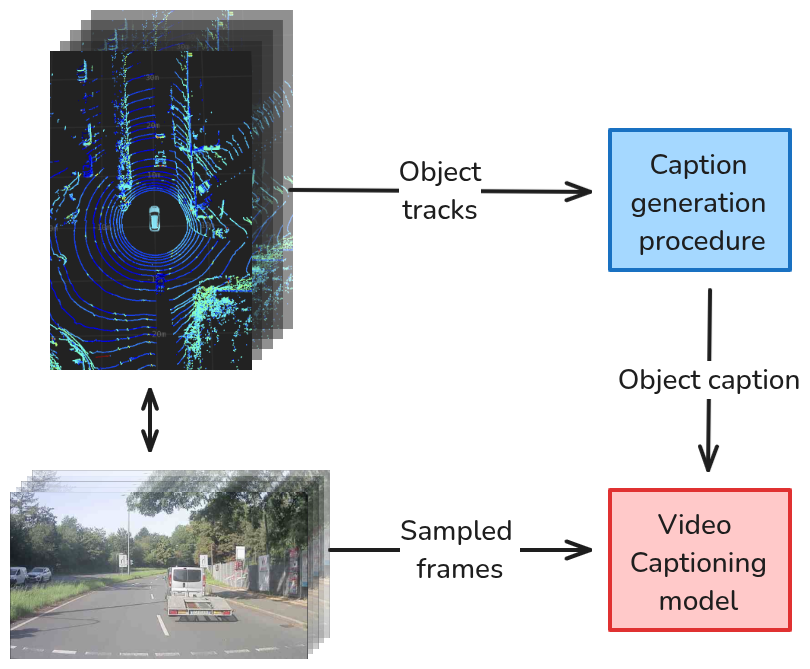}
    \caption{\textbf{LiDAR-based caption generation and video captioning model training.} This figure shows the overall workflow of our approach. Object tracks extracted from raw LiDAR data using a SOTA 3D detector and tracker are used to identify event boundaries and generate object-level captions. The object caption and associated front camera RGB images (using event boundaries) are used to train a video captioning model (multi-modal masked language modeling).}
    \label{fig: system design}
\end{figure}

Video captioning \cite{kuo2023mammutsimplearchitecturejoint,he2023vlabenhancingvideolanguage,chen2023valorvisionaudiolanguageomniperceptionpretraining} and foundation models \cite{wang2022internvideo,wang2024internvideo2,cheng2024videollama} have shown strong semantic capabilities on open-world datasets. However, their performance in street scenes still requires improvement. Specifically, off-the-shelf models \cite{lin2022swinbert,wang2022internvideo,cheng2024videollama} lack precision in capturing detailed scene attributes (e.g., identifying the lane of a car) and dynamics (e.g., describing vehicle movements relative to the host).

Recently, there has been a surge in the creation of language datasets for street scenes \cite{marculingoqa,qian2024nuscenes,inoue2024nuscenes,wu2023language,arai2024covla} focusing on question answering and driving instructions, facilitating the exploration of video captioning models within this domain. These datasets have been annotated by humans, partially assisted by large language models (LLMs). In addition to that, NuScenes-QA \cite{qian2024nuscenes} utilizes Light Detection and Ranging (LiDAR) 3D bounding box information to create text pairs for question answering, allowing to pose question like ``Are there any standing pedestrians to the front right of the stopped bus?''. 

None of the above datasets provide captions with precise scene details or descriptions of temporal dynamics. To address this, a systematic framework is needed to generate rich, temporally grounded descriptions from multimodal driving data. For instance, a caption like ``a car is overtaking the ego vehicle from the left and then turning right'' captures both spatial relations and temporal semantics, details often absent in existing datasets. We address this lack of data by introducing a generic procedure to caption street scene videos using LiDAR point clouds. From a given video sequence with corresponding LiDAR sweeps, we infer the locations of various traffic participants over the course of a video using a strong object detector (LiDAR) and tracker model. The resulting time series of locations/trajectories are combined with lane information and then aggregated in order to create captions in a rule-based manner. These captions provide fine-grained information about the traffic participant's actions. The information of the ego vehicle and all other traffic participants are accumulated, providing a rich and detailed description of the temporal dynamics of traffic participants. Using the resulting captions on a proprietary dataset as well as on NuScenes \cite{caesar2020nuscenes} and Waymo \cite{sun2020scalability}, we train the SwinBERT \cite{lin2022swinbert} video captioning model to predict these temporally grounded descriptions. 

Crucially, LiDAR data is employed exclusively during caption generation and is entirely absent from model training or inference. Our approach is broadly applicable, seamlessly extending to any dataset with corresponding LiDAR sweeps and camera videos. Leveraging off-the-shelf LiDAR object trackers, this method eliminates reliance on human annotations, enabling scalable deployment across extensive datasets. We summarize our contributions as follows:
\begin{itemize}
  \item We introduce the first fully scalable and generic procedure for generating pseudo-ground truth captions that provide fine-grained temporal descriptions of traffic participants in street scenes using a rule-based system.
  \item We train SwinBERT \cite{lin2022swinbert} using the generated video captions. To assess the improvement in the model's temporal understanding, we conducted several studies, including a video retrieval task leveraging the hidden states of SwinBERT to identify scenes exhibiting similar temporal dynamics, such as object maneuvers.
  \item We propose a novel metric, Visual Bias Measure (VBM), designed to quantify the extent of visual bias present in a model’s embedding space. This metric serves as an indirect indicator of the model’s temporal understanding, with lower visual bias suggesting stronger sensitivity to temporal dynamics. Using our custom VBM metric, we show that training a model using our temporal captions helps to mitigate the inherent visual bias present in state-of-the-art (SOTA) models. We also highlight the generalization capability of our methodology by showing reduced VBM consistently across 2 public datasets, even with a different camera setup compared to the training dataset.
\end{itemize}

\section{Related work}
\label{sec:related work}
\textbf{Language-based driving dataset}: Most large open-source public datasets \cite{caesar2020nuscenes, liao2022kitti, sun2020scalability} contain multi-modal data but often lack language annotations such as scene descriptions. Manual annotation requires the development of an extensive annotation schema and remains highly dependent on individual annotators, leading to variability. This challenge has contributed to the growing use of LLMs to refine text descriptions, producing more human-like language while significantly reducing costs.

For instance, NuScenes QA \cite{qian2024nuscenes} generates scene graphs using LiDAR bounding box data to represent scenes. By employing template-based question construction, they create 460k question-answer pairs (one-word answers) from 34k driving scenes. Similarly, NuScenes MQA \cite{inoue2024nuscenes} focuses on generating question-answer datasets using NuScenes ground truth annotations, but it differs from NuScenes QA by producing full-sentence answers with GPT-4 and human reviewers. NuPrompt \cite{wu2023language}, on the other hand, uses object location data from multi-view cameras and manually assigns attributes to generate basic object descriptions within scenes. GPT-3.5 is then employed to craft human-like prompts, which are subsequently reviewed by human annotators, resulting in around 85k prompts from 850 videos. 

Our approach seeks to fully automate the annotation process by replacing human involvement with a rule-based system to describe the actions of road participants, and utilizing a 3D LiDAR detector and tracker instead of manual annotation. Although the template-based approach constrains linguistic variety, the combinatorial design of templates still yields a substantial range of possible captions, as described in \cref{sec:data generation}.

\textbf{Video understanding models}: Research on vision-language models has advanced significantly, particularly with the introduction of Vision Transformers \cite{alexey2020image} and their popular variants like \cite{liu2021swin, touvron2021training, bertasius2021space, arnab2021vivit}. A key distinction in current model architectures is whether the vision and language models are trained jointly or separately, with joint training proving more effective. Numerous studies have focused on developing both task-specific \cite{zhao2022tuber, wang2021end} and foundation models \cite{zhang2023video, wang2022internvideo, wang2024internvideo2, lin2023video} for video understanding in open-world scenarios. However, adapting these models to autonomous driving remains an emerging area.
The Action Aware Driving Captioning Transformer \cite{jin2023adapt} adapts SwinBERT \cite{lin2022swinbert} for driving scenarios using the BDD-X dataset \cite{kim2018textual}, but its focus is limited to causal understanding of driver actions rather than describing the actions of all objects in the scene. The model architecture proposed by NuPrompt \cite{wu2023language} is similar to ours in that it utilizes only camera images (from multiple views) as input, along with a prompt, to output bounding boxes. However, our approach differs by using only front camera images and generating captions for all tracked objects within a predefined neighborhood. In our method, caption generation is fully automated, and the training is self-supervised. %, as outlined in \cite{lin2022swinbert}. 
This enables scalable data generation and model training.
%%%%%%%%%%%%%%%%%%%%%%%%%%%%%%%%%%%%%%%%%%%%%%%%%%%%%%%%%%%%%%%%%%%%%%%%
\section{Methodology}
\label{sec:data generation}
In this section, we introduce our cross-modality-based method for generating object-specific captions describing traffic dynamics to train a video captioning model. In \cref{sec: video captioning procedure}, we detail the automatic generation of captions from LiDAR object tracks using generic templates, divided into descriptions of host vehicle actions and actions of other traffic participants. Thereafter, in \cref{sec: model training}, we outline the adaptation of these captions to train a SwinBERT model. To assess improvements in temporal understanding and mitigation of visual bias, we adopt the Vision Transformer-Base (ViT-B) architecture and Masked Language Modeling (MLM) strategy proposed by \cite{lin2022swinbert}. We introduce a custom metric (detailed later) to quantify visual bias in the model’s embeddings, offering insight into its temporal reasoning. Additionally, we evaluate the effect of using frames augmented with automatically generated object-level masks (from point clouds) on caption quality.
 
 \subsection{LiDAR-based captioning procedure}
 \label{sec: video captioning procedure}

\begin{figure*}[t]
    \centering
    \includegraphics[width=0.9\textwidth]{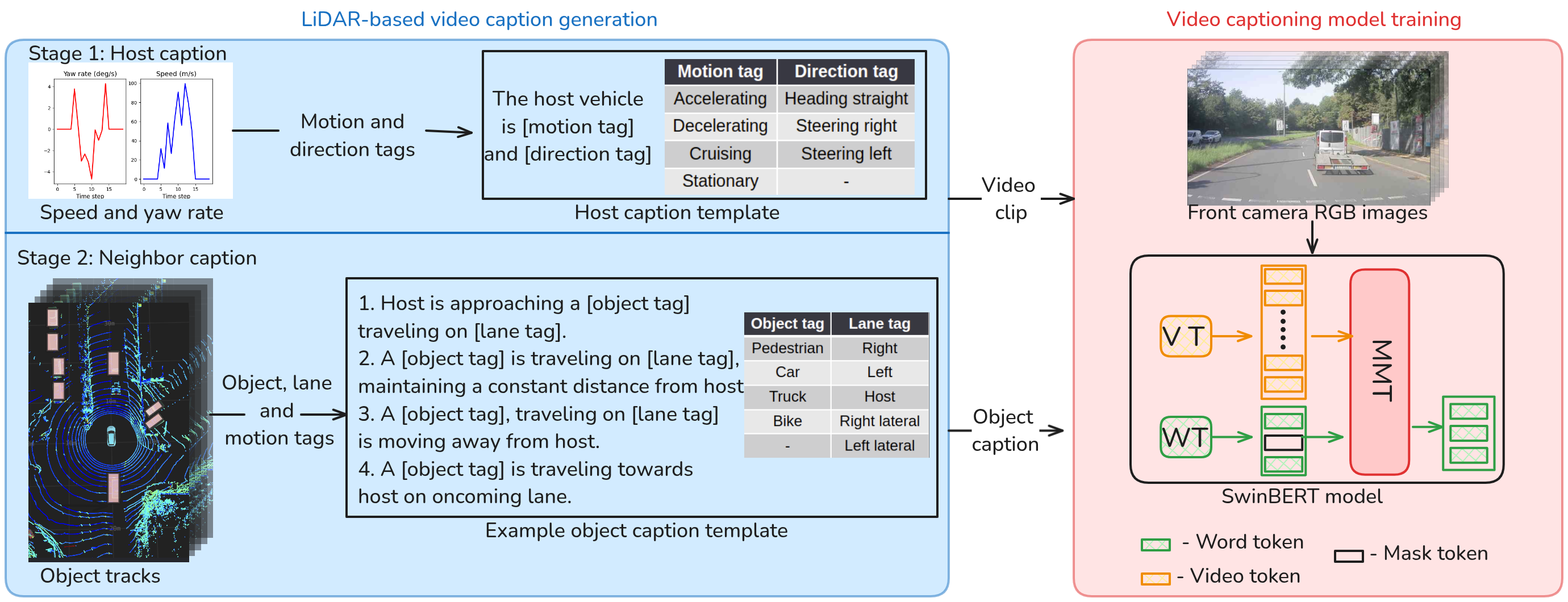}
    \caption{\textbf{LiDAR-based caption generation and video captioning model training.} This figure illustrates our overall caption generation procedure and the training process for the video captioning model. Using the host vehicle's sensor data (speed and yaw rate) and object tracks from LiDAR data, we generate the object, lane and motion tags. These tags are used in the corresponding placeholders of the template sentence to construct the host and object captions. Finally, the object captions with the front camera RGB images are used to train a video captioning model (SwinBERT)through Masked Language Modeling. VT, WT, and MMT stand for Vision transformer, Word tokenizer, and Multi-modal transformer, respectively.}
    \label{fig: methodology system design}
\end{figure*}

Our proposed data generation methodology is fully automated, requiring no human involvement. Captions are generated using template sentences with placeholders (\cref{fig: methodology system design}), populated from a predefined set of options. These placeholders capture temporal dynamics, including lane information, travel direction, and relative distance changes. Combining these three attributes yields diverse captions that accurately represent various traffic scenarios, e.g., ``Host is approaching a car traveling on the right lane. It is now on the right lateral lane and moving away from host.''
The combinatorial nature of host captions and an example of object captions are illustrated in \cref{fig: methodology system design}. Notably, unlike most video captioning datasets, our captions can span multiple sentences, referred to as caption length. Generally, more complex object actions produce longer captions. Further details on our caption generation process as well was the relationship between caption length and action complexity are discussed later in this section.

The LiDAR-based captioning process consists of two stages. In the first stage, the video is split into smaller clips that each capture a single, continuous action by the host vehicle, using sensor data from the host. This segmentation is essential, as all interactions with other road users are described relative to the host vehicle.
In the second stage, the captions focus on the interactions between the host and other traffic participants. First, tracks of these participants (over time) are derived from LiDAR data using SOTA 3D bounding box detectors and trackers, hereafter referred to simply as tracks. These tracks are then fed into a rule-based system to generate generic captions describing the participants’ dynamics, as illustrated in \cref{fig: methodology system design}.

\textbf{Host caption generation.} We begin by analyzing the host vehicle’s velocity and yaw rate at every timestamp (see \cref{fig: methodology system design}). Comparing these values to predefined thresholds yields two tags, motion (acceleration, deceleration, cruising, or stationary) and direction (left, right, or straight). (refer to supplementary material for more details.)

Frames sharing the same motion–direction pair are grouped into segments that capture a single, consistent action (e.g., “decelerating and steering left”). Only segments 8 to 300 frames long are kept, avoiding clips that are too brief or overly long. Each segment receives a host caption generated from a template containing the motion and direction placeholders. Since acceleration, deceleration, and cruising motion tags combine with three direction tags and “stationary” tag combines only with “straight”, the system can produce up to ten distinct host captions. These captions supply the context for object-level captions generated in the next stage.

\textbf{Neighbor caption generation.} In stage two, we generate object captions for each segmented clip. Object tracks are extracted from raw LiDAR via a SOTA 3D detection and tracking pipeline (see \cref{fig: methodology system design}). We then filter for relevance: (1) retain only moving cars, trucks, bicycles, and pedestrians, and (2) keep those within a rectangular neighborhood directly ahead of the host inside the front-camera field of view. The threshold logic used to identify objects in this neighborhood is illustrated by \cref{eq: neighborhood}, while further details on filtering out stationary objects is provided in the supplementary material.

\begin{equation}
    \label{eq: neighborhood}
    \text{Neighbor} = 
    \begin{cases}
    \text{Yes}, & \text{if } 0 < x < T_x \text{ and} -T_y < y < T_y\\
    \text{No}, & \text{otherwise }
    \end{cases}
\end{equation}
where, $x$ and $y$ are the distances of the object's center from the host vehicle w.r.t.\ heading direction ($x$) and perpendicular direction ($y$), respectively, and $T_x$ and $T_y$ denote threshold values on those respective distances. In essence, \Cref{eq: neighborhood} defines the neighborhood, a rectangular region in front of the host vehicle with length $T_x$ and width $2 \times T_y$, and objects only in this neighborhood are considered for the neighbor caption generation. To ensure objects are visible to the front camera, we also project their 3D bounding boxes onto the 2D image plane, while accounting for lens distortion.

In summary, object captions are generated only for moving neighbors that fall within a defined spatial region in front of the host and are visible to the front camera, ensuring that the generated captions focus on relevant and temporal dynamics of the scene.

At every LiDAR timestamp, we assign lane and motion tags to each neighbor object, forming a time series that describes the object’s behavior throughout the clip. Lane tags are determined based on two key factors: (1) the yaw angle of the neighbor relative to the host vehicle, and (2) the lateral position of the neighbor with respect to the host (i.e., its side-to-side location in the coordinate frame of the host). Using predefined thresholds on these values, each object is assigned one of the following lane tags: \texttt{right}, \texttt{left}, \texttt{host}, \texttt{oncoming}, \texttt{left lateral}, or \texttt{right lateral lane}.

Motion tags are assigned by analyzing how the relative distance between the neighbor and the host changes between consecutive frames. Depending on whether the object is getting closer, moving away, staying at a constant distance, or not moving at all, it receives one of the following tags: \texttt{approach}, \texttt{away}, \texttt{constant distance}, or \texttt{stationary}. \Cref{table:neighbor caption-placeholder label} provides example values for the object tag, lane tag, and motion tag. These three tags are combined at each frame to form a structured representation we refer to as a \emph{concatenated tag} (e.g., \texttt{car-host-away}).

To avoid repetitive captions and capture only meaningful behavior changes, we compress the time series of concatenated tags into a \emph{unified sequence} by keeping only the distinct combinations of concatenated tags in the order they appear. This provides a high-level summary of the object’s activity throughout the clip. For example: Suppose a neighbor object, tagged as a \texttt{car}, receives the following concatenated tags over time: (\texttt{host-away}, \texttt{host-away}, $\ldots$, \texttt{right-away}, \texttt{right-away}, $\ldots$, \texttt{right-away}). Then, the unified tag sequence becomes: (\texttt{car-host-away}, \texttt{car-right-away}). Each entry in this sequence corresponds to a distinct behavior of the object and maps to one sentence in the final caption.

\begin{table}[t]
  \caption{Placeholder values for neighbor attributes. A combination of the object, lane, and motion tag generates the caption for each moving object in the neighborhood. This results in a large pool of possible captions.}
  \centering
  \begin{tabular}{ccc}
    \toprule
    Object tag & Lane tag & Motion tag \\
    \midrule
    Pedestrian & Right & Approach \\
    Car & Left & Away \\
    Truck & Host & Constant \\
    Bike & Oncoming & Stationary \\
    - & Left Lateral & - \\
    - & Right Lateral & - \\
    \bottomrule
  \end{tabular}
  \label{table:neighbor caption-placeholder label}
\end{table}

Once the unified tag sequence is obtained for a moving neighbor object, the final neighbor caption is generated using sentence templates. Each unique concatenated tag in the unified tag sequence (e.g., \texttt{car-host-away}) corresponds to one sentence in the caption. These templates help convert structured tag information into natural language descriptions of the object's behavior over time. We have a set of predefined sentence templates that vary depending on: the object tag, motion tag and position of the concatenated tag in the unified tag sequence. These templates allow us to construct fluent and informative sentences while maintaining consistency in how different actions are described.

The first tag in the unified sequence is treated as the object’s initial state in the segment. For this case, the template focuses on introducing the object type (car), its lane position (host) and motion information (moving away from host). For example, the unified tag \texttt{car-host-away} would yield the caption: \texttt{A car, traveling on the host lane, is moving away from the host}. This defines the scene for the object’s starting behavior. For each subsequent tag, the template is more nuanced, it captures not only the object’s current state, but also how it has changed from the previous state. This helps describe transitions in behavior, such as lane changes or shifts in motion. These subsequent sentences typically encapsulate: (1) a reference to the continued presence of the object, and (2) a description of what changed. For example, If the next tag in the sequence is \texttt{car-right-away}, and the previous one was \texttt{car-host-away}, the generated sentence would be: \texttt{It continues to move away from the host, but is now on the right lane.}
This formulation highlights that the object is still moving away but has changed its lane from host to right. The final caption is assembled by concatenating all the generated sentences in the sequence, maintaining their order. The number of sentences directly reflects the number of unique behavioral changes the object underwent during the segment.

The first sentence of each neighbor caption is generated using a dedicated set of template sentences, specifically designed for the motion tags \texttt{approach}, \texttt{constant}, and \texttt{away}. There is also a special case for the \texttt{oncoming} lane tag, which is only compatible with the \texttt{approach} motion tag. These initial sentence templates are shown in \cref{fig: methodology system design} and account for 63 unique variations, forming the ``pool'' of possible first sentences. For all subsequent sentences in multi-sentence captions, we use a separate set of templates outlined in \cref{table:neighbor caption-template filler,table:neighbor caption-templates}. These templates handle transitions in behavior (e.g., lane changes or motion shifts) and come in three distinct styles, resulting in 78 unique sentence options for follow-up descriptions.

When combining templates to create full captions, the number of possibilities grows rapidly. For example, a caption consisting of two sentences can be formed by selecting one option from the 63 first-sentence pool and one from the 78 follow-up sentence pool, giving: $63 \times 78=4,\!914$ unique two-sentence captions. More generally, the total number of possible captions of length 
n can be calculated using the formula: $|A| \cdot \frac{|B|!}{(|B|-(n-1))!}$ , where $A$ and $B$ are sets of possible captions for the first sentence (\cref{fig: methodology system design}) and follow-up sentences (\cref{table:neighbor caption-template filler,table:neighbor caption-templates}).

\begin{table}[t]
  \caption{Table showcasing the different filler values used in the template sentences, determined by both the motion tag and the host action.}
  \centering
  \begin{tabular}{ccc}
    \toprule
    Motion tag & Host action & Motion filler \\
    \midrule
    Away & N/A & Moving/move away from host \\
    Constant & N/A & \makecell{Maintaining/maintain a constant \\distance from host} \\
    Stationary & N/A & Stationary \\
    Approach & Stationary & Approaching/approach host \\
    Approach & Non-stationary & Distance from host is reducing/reduce \\
    \bottomrule
  \end{tabular}
  \label{table:neighbor caption-template filler}
\end{table}

\begin{table}[t]
  \caption{Template sentences for neighbor caption generation are selected based on the lane tag and the motion filler (derived from the motion tag, as presented in \cref{table:neighbor caption-template filler}). The templates vary depending on whether the neighbor’s motion and/or lane change between consecutive frames. In this table, the first column indicates whether the motion tag differs between two consecutive frames, while the second column indicates whether the lane tag changes}
  \centering
  \begin{tabular}{ccc}
    \toprule
    \makecell{Differs in\\motion tag} & \makecell{Differs in\\motion tag} & Template \\
    \midrule
    Yes & No & \makecell{It/He/She continues to be on the \textbf{[lane tag]} \\lane/sidewalk but is now \textbf{[motion filler]}.}\\
    No & Yes & \makecell{It/He/She continues to \textbf{[motion filler]} \\but is now on the \textbf{[lane tag]} lane/sidewalk.}\\
    Yes & Yes & \makecell{It/He/She is now on the \textbf{[lane tag]} \\lane/sidewalk and is \textbf{[motion filler]}.}\\
    \bottomrule
  \end{tabular}
  \label{table:neighbor caption-templates}
\end{table}

 \subsection{Video Captioning model}
 \label{sec: model training}
Using the video captions constructed by our LiDAR-based procedure, we train the SwinBERT video captioning model. This model processes front camera RGB images to predict object-level captions. We chose SwinBERT since our captions are template-based---focusing on temporal dynamics rather than linguistic variety---making large generative language models unnecessary. Resource constraints also influenced our preference for a smaller yet efficient captioning model like SwinBERT.

We limit our primary experiments to a subset of the dataset consisting of clips with captions of length 1. Additionally, we benchmark SwinBERT (trained on our captions constructed from LiDAR) against two video-language models InternVideo\cite{wang2022internvideo} and ViCLIP \cite{wang2023internvid}, an image-language model CLIP \cite{radford2021learning}, and an object classifier VGG19 \cite{simonyan2014very} to evaluate improvements in temporal understanding, as discussed later in this section. The benefits of LiDAR-based caption supervision are both scalable and cost-effective, making the overall procedure adaptable to any multi-modality training framework, while enhancing the temporal understanding of models by reducing the visual bias. Additionally, we also conduct ablation studies on the model’s zero-shot performance---focusing on more complex object behaviors and edge-case scenarios---to assess its generalization capabilities.

\textbf{LiDAR-based mask generation.} Building on the insights from \cite{fukuzawa2025can}, demonstrating the benefits of frame masking in reducing visual bias, we incorporate an automated LiDAR-based object masking technique into our procedure. This approach helps the model’s to focus on relevant scene elements by providing additional visual cues that explicitly link objects to their corresponding captions. Unlike open-domain video captioning datasets, where a frame often centers around a single subject or event, street scenes are more complex, typically featuring multiple objects and concurrent activities. In such scenarios, object-level masking becomes especially valuable, as it helps isolate the specific visual regions associated with each caption.

To create these object-level masks, we follow a 3-step pipeline: (1) Point cloud extraction: For each target object, we first extract the corresponding 3D point cloud enclosed within its bounding box. (2) Projection to image plane: The extrtacted 3D points are projected onto the 2D image plane of the camera using an affine transformation, which is derived from the LiDAR-to-camera calibration and camera's intrinsic parameters. This step ensures spatial alignment between LiDAR and camera views. (3) Mask contour construction: Once projected onto the image plane, the 2D points are used to construct the mask boundary as a convex hull. We apply Andrew’s Monotone Chain algorithm \cite{andrew1979another} to generate a convex polygon that tightly wraps around the projected points, forming the mask's contour.

\begin{figure}[t]
    \centering
    \includegraphics[width=0.49\textwidth]{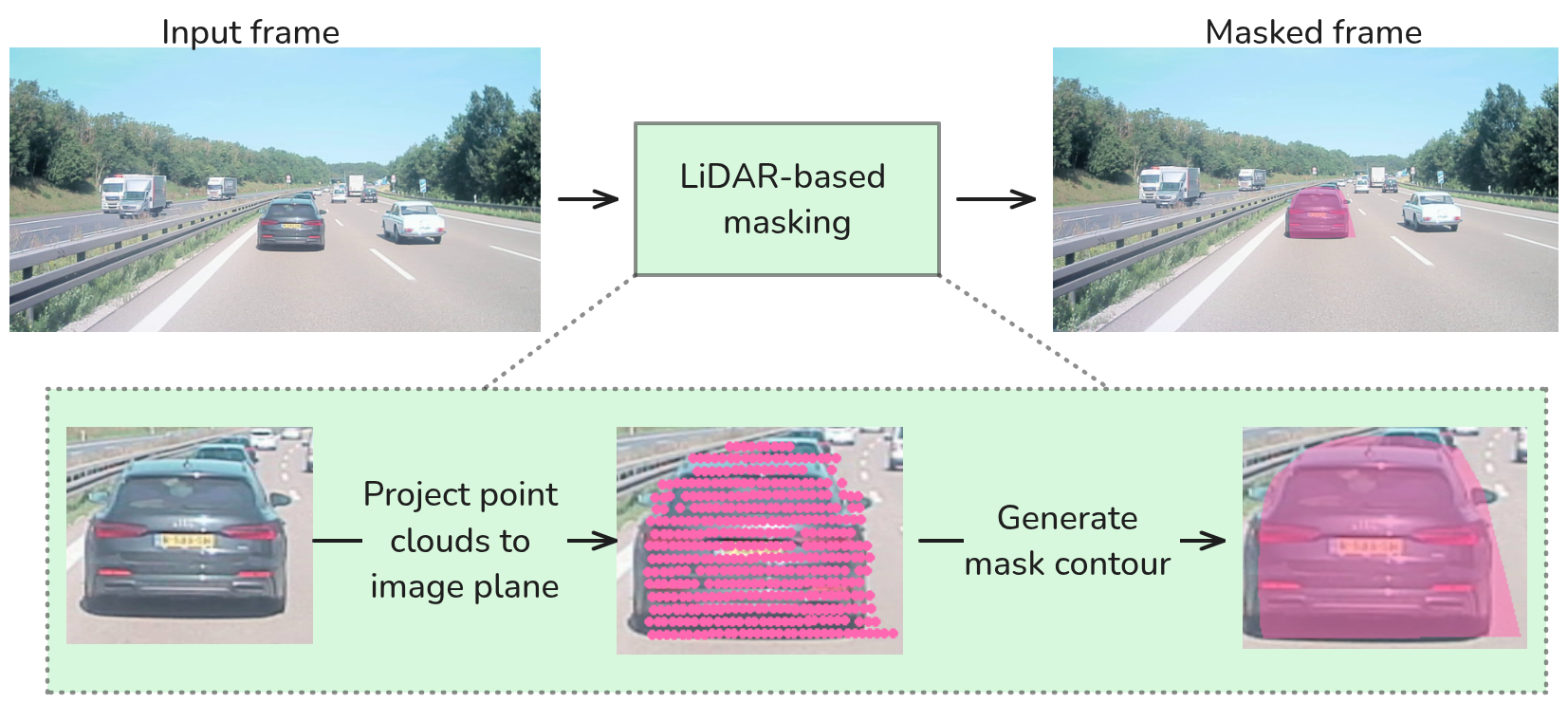}
    \caption{\textbf{LiDAR-based object mask generation.} This figure highlights our mask generation procedure. For a given input frame, the point clouds corresponding to the object of interest (black car) are first projected onto the front-camera image plane, forming a dense 2-D scatter (center image in the green box). Andrew’s Monotone Chain convex-hull algorithm then traces a tight contour around these projected points, which is rasterised into a mask that cleanly outlines the object (right image in the green box).}
    \label{fig: methodology masking}
\end{figure}

The result is a clean, object-specific mask that highlights only the region associated with the object of interest, minimizing distractions from unrelated parts of the scene. An illustration of a masked frame, along with the intermediate steps of the mask generation process, is shown in \cref{fig: methodology masking}.

\subsection{Benchmarking}
 \label{sec: model benchmarking}
To evaluate the temporal understanding embedded in the model’s video representations, we use a video-to-video retrieval framework and compare the model’s performance with other SOTA approaches. For each query video, we compute the cosine similarity between its embedding and those of all other videos in the test set. The top-k most similar videos (nearest neighbors) are identified based on these similarity scores. To assess retrieval quality, we compute the mean $\mathrm{BLEU4}$ score \cite{papineni2002bleu} between the ground-truth captions of the query video and those of its top-k retrieved neighbors. Higher $\mathrm{BLEU4}$ scores indicate stronger alignment between the query and its retrieved neighbors in terms of semantic content.

However, videos recorded within a short time window (e.g., within minutes of each other) often capture the same or similar surroundings, which may lead the model to rely heavily on visual appearance rather than true temporal understanding. To account for this, we introduce a constrained retrieval setting. In this setting, we exclude all videos that were recorded within a predefined time window around the query video, effectively forcing the model to retrieve videos that were recorded at different times, likely in different scenes or contexts. The goal is to test whether the model can retrieve semantically and temporally relevant videos without depending on low-level visual similarity.

We compute $\mathrm{BLEU4}$ scores again under this constraint, and compare them with the original (unconstrained) retrieval results. To quantify the impact of visual bias, we define the Visual Bias Measure (VBM) as the relative percentage drop in $\mathrm{BLEU4}$ score due to this temporal exclusion:
\begin{equation}
    \label{eq: VBM}
    \mathrm{VBM} = 100 \cdot \frac{B_k - C_k}{B_k}    
\end{equation}

where, $B_k$ and $C_k$ are the $\mathrm{BLEU4}$ scores computed using the original top-$k$ neighbors (no time constraints) and using only time-separated neighbors in the constrained setting, respectively. A higher VBM score indicates a larger drop in retrieval quality when nearby-in-time videos are excluded, suggesting the model is overly reliant on visual similarity, i.e., it exhibits strong visual bias and limited temporal generalization.
%%%%%%%%%%%%%%%%%%%%%%%%%%%%%%%%%%%%%%%%%%%%%%%%%%%%%%%%%%%%%%%%%%%%%%%%
\section{Experiments}
In \cref{sec: implementation}, we outline the experimental setup and key implementation details of our caption generation framework. \Cref{sec: inference} presents a quantitative evaluation of captioning performance and generalization through zero-shot predictions on two large-scale autonomous driving datasets: Waymo and NuScenes. We benchmark our approach against several SOTA methods to provide a meaningful comparison. \Cref{sec: vbm evaluation} introduces the Visual Bias Measure (VBM) to quantify improvements in temporal representation and reduction in visual bias within the model’s embedding space. We further assess the model’s generalization to more complex descriptions by evaluating zero-shot performance on two-sentence captions, despite training exclusively on single-sentence captions. Finally, \cref{sec: qualitative} offers a qualitative analysis of the learned embedding space, using UMAP projections and visual examples to illustrate how temporal patterns are encoded.

\subsection{Experimental setup}
\label{sec: implementation}
\begin{figure}[t]
    \centering
    \includegraphics[width=0.35\textwidth]{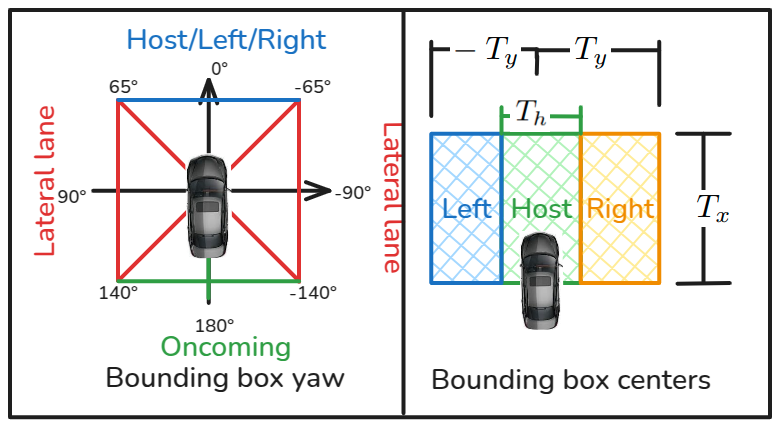}
    \caption{\textbf{Thresholds for lane tag.} This figure illustrates the threshold values used in our neighbor captioning process. On the left, we depict the yaw angle thresholds applied to bounding boxes for lane tagging. On the right, we show the thresholds used for determining lane tags based on bounding box center positions.}
    \label{fig: neighbor lane tag}
\end{figure}

We apply our captioning procedure, detailed in \cref{sec:data generation}, to a proprietary dataset and two public datasets (Waymo and NuScenes). For all subsequent experiments, we train the SwinBERT model exclusively on single-sentence object captions, consistent with the evaluation setup used in prior work on the SwinBERT architecture. \Cref{fig: neighbor lane tag} and \cref{table: neighbor threshold parameters} illustrate the thresholds used for generating neighbor captions. 

\begin{table}[t]
    \caption{\textbf{Threshold values for neighbor caption construction.} This table lists the threshold values used to assign lane and motion tags to neighbors. Parameters $T_x$, $T_y$ and $T_h$ are depicted in \cref{fig: neighbor lane tag}. $T_s$ is the threshold parameter for identifying non-stationary objects, and $T_m$ is for the assignment of motion tags. For more details on these parameters, please refer to the supplementary material.}
    \centering
    \begin{tabular}{ccc}
        \toprule
        \textbf{\thead{Threshold\ Parameter}} & \textbf{\thead{Neighbor}} & \textbf{\thead{Threshold\ Value}} \\
        \midrule
        \multirow{2}{*}{$T_x$} & Pedestrian & 30 \\
        & Car/Truck/Bike & 40 \\
        \midrule
        \multirow{2}{*}{$T_y$} & Pedestrian & 15 \\
        & Car/Truck/Bike & 20 \\
        \midrule
        \multirow{2}{*}{$T_s$} & Pedestrian/Bike & 0.1 \\
        & Car/Truck & 0.15 \\
        \midrule
        \multirow{2}{*}{$T_m$} & Pedestrian & 0.01 \\
        & Car/Truck/Bike & 0.1 \\
        \midrule
        $T_h$ & Any & 2.5 \\
        \bottomrule
    \end{tabular}
    \label{table: neighbor threshold parameters}
\end{table}

We train the SwinBERT model on the selected subset of our proprietary dataset, captions with only one sentence, comprising approximately 50k video-caption pairs (only front camera video recording). To further explore the model’s performance, we also experiment with mask-augmented video frames and analyze their impact on captioning quality. All investigations are conducted with model trained by uniformly sampling 8 frames from the video clips and the model training takes around 4 days on a single A100 NVIDIA GPU. 

\subsection{Caption prediction}
\label{sec: inference}
We train the SwinBERT model exclusively on the proprietary video dataset and evaluate its captioning quality on the test splits of the proprietary dataset, the entire Waymo, and the entire NuScenes datasets. The evaluation is conducted using a diverse set of natural language processing (NLP) performance metrics, $\mathrm{BLEU4}$, $\mathrm{CIDEr}$, and $\mathrm{SPICE}$, as detailed in \cref{res:nlp evaluation}.

The quantitative results in \cref{res:nlp evaluation} highlight the strong generalization ability of our model, which was trained solely on the proprietary dataset. Despite this limited training scope, our model consistently outperforms all baselines, InternVideo, ViCLIP, and CLIP, on both seen and unseen datasets. On the proprietary dataset, our model outperforms the next best-performing methods by a margin of 3× in $\mathrm{BLEU4}$, 3.2$\times$ in $\mathrm{CIDEr}$, and 2$\times$ in $\mathrm{SPICE}$, demonstrating substantial gains across all evaluation metrics. On the more challenging Waymo and NuScenes datasets---which feature different camera configurations---our model continues to show strong generalization, achieving at least 1.6$\times$ higher  $\mathrm{BLEU4}$, 2$\times$ higher $\mathrm{CIDEr}$, and 1.9$\times$ higher $\mathrm{SPICE}$ scores compared to the respective next best models.

These results are even more striking given that the baseline models were not required to generate captions; instead, they were evaluated by selecting the best match from a list of all possible captions, a significantly easier task. Despite this advantage, the baselines fail to match the performance of our model, underscoring the effectiveness of our temporal captions.

\begin{table*}[t]
  \caption{\textbf{Evaluation of Caption prediction.} The captioning quality of our model, assessed through various NLP metrics. Trained exclusively on our proprietary dataset, the model’s zero-shot inference results are shown for the Waymo and NuScenes datasets. }
  \centering
  \begin{tabular}{ccccccccccccccc}
    \toprule  
    && \multicolumn{3}{c}{\textbf{Ours}} & \multicolumn{3}{c}{\textbf{InternVideo}} & \multicolumn{3}{c}{\textbf{ViCLIP}} & \multicolumn{3}{c}{\textbf{CLIP}}\\
    \cmidrule(lr){3-5} \cmidrule(lr){6-8} \cmidrule(lr){9-11} \cmidrule(lr){12-14}
    \textbf{\thead{Dataset}} & \textbf{\thead{Object\\mask}} & \textbf{\thead{BLEU4}} & \textbf{\thead{CIDEr}} & \textbf{\thead{SPICE}} & \textbf{\thead{BLEU4}} & \textbf{\thead{CIDEr}} & \textbf{\thead{SPICE}} & \textbf{\thead{BLEU4}} & \textbf{\thead{CIDEr}} & \textbf{\thead{SPICE}}
    & \textbf{\thead{BLEU4}} & \textbf{\thead{CIDEr}} & \textbf{\thead{SPICE}}\\
    \midrule
    \multirow{2}{*}{Proprietary} & Yes & \textbf{0.80} & \textbf{4.50} & \textbf{0.88} & 0.13 & 0.66 & 0.39 & 0.22 & 1.21 & 0.41 & 0.16 & 0.74 & 0.39\\
    & No & 0.70 & 3.83 & 0.80 & 0.14 & 0.72 & 0.40 & 0.23 & 1.41 & 0.45 & 0.15 & 0.66 & 0.37\\
    \midrule
    \multirow{2}{*}{Waymo} & Yes & \textbf{0.67} & \textbf{3.84} & \textbf{0.77} & 0.15 & 0.80 & 0.30 & 0.22 & 1.42 & 0.36 & 0.18 & 1.04 & 0.41\\
    & No & 0.56 & 3.15 & 0.69 & 0.15 & 0.79 & 0.29 & 0.19 & 1.21 & 0.34 & 0.17 & 0.94 & 0.38\\
    \midrule
    \multirow{2}{*}{NuScenes} & Yes & 0.37 & 1.31 & \textbf{0.52} & 0.09 & 0.39 & 0.30 & 0.13 & 0.65 & 0.26 & 0.07 & 0.30 & 0.21\\
    & No & \textbf{0.43} & \textbf{1.37} & 0.49 & 0.10 & 0.38 & 0.29 & 0.13 & 0.74 & 0.27 & 0.06 & 0.29 & 0.21\\
    \bottomrule
  \end{tabular}
  \label{res:nlp evaluation}
\end{table*}

\subsection{Visual bias evaluation}
\label{sec: vbm evaluation}
To quantify the reduction of visual bias (improvement in temporal understanding) achieved by our captions, we compare the SwinBERT model trained on our proprietary dataset against three foundational multimodal LLMs (zero-shot prediction) using our custom metric, VBM. Two of these models, InternVideo\cite{wang2022internvideo} and VideoCLIP\cite{wang2023internvid}, were originally trained on image and video datasets, while the third, CLIP \cite{radford2021learning}, was trained exclusively on images. Additionally, VGG19 \cite{simonyan2014very}, a model trained as an object classifier, is used as a fourth benchmark. The Waymo and NuScenes datasets are included in the VBM evaluation to assess the generalization of our data generation and training methodology.

\begin{figure}[t]
    \centering
    \includegraphics[width=0.4\textwidth]{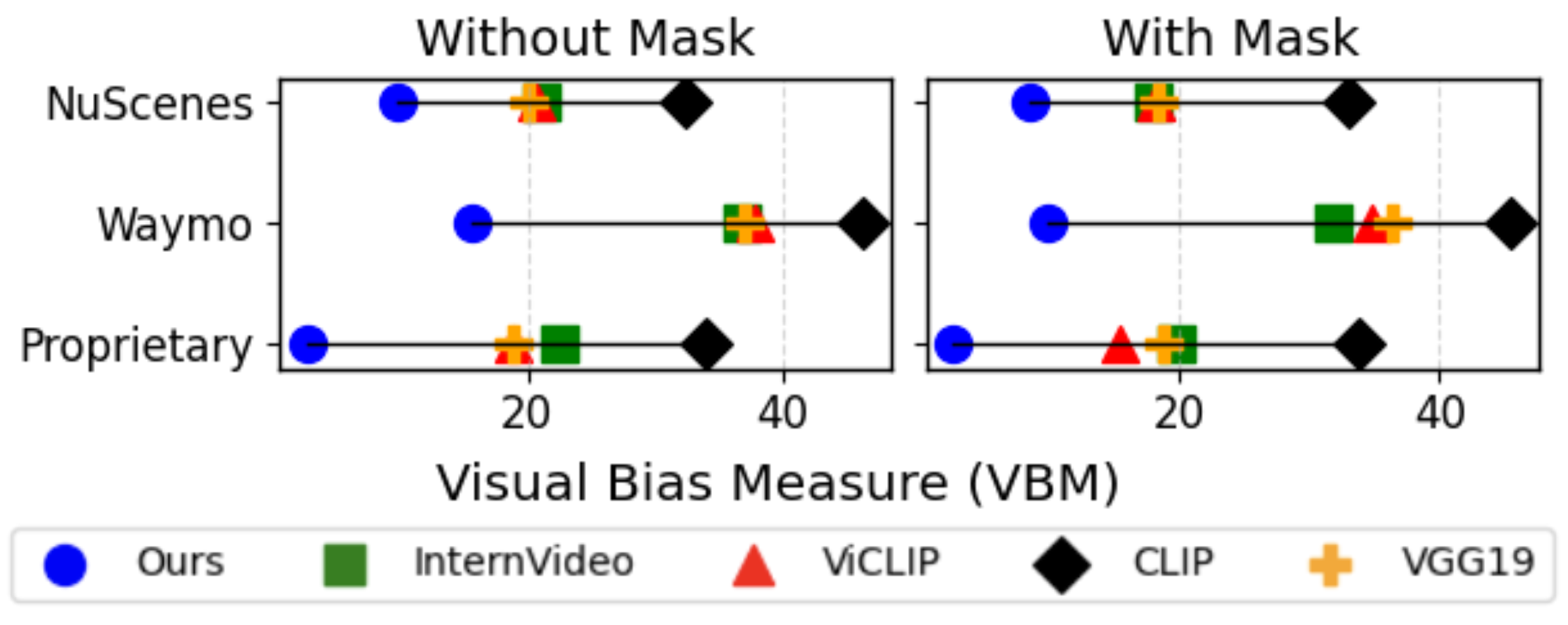}
    \caption{\textbf{Visual Bias Measure (VBM) based comparison.} A comparison of our model, trained using our proprietary dataset, with several SOTA models via a video-to-video retrieval approach, evaluated with our custom VBM metric across three datasets. Left and right plots illustrate the benchmarking results using mask-augmented frames and raw frames, respectively. A lower VBM score indicates that the model's embeddings are less influenced by visual bias present in the video.}
    \label{fig: retrieval bleu reduction}
\end{figure}

\Cref{fig: retrieval bleu reduction} shows the reduction in VBM achieved using our captions. Despite InternVideo and VideoCLIP being designed to enhance temporal understanding (e.g., through the VideoMAE training strategy), our approach consistently exhibits the lowest VBM across all datasets. On our proprietary dataset without mask augmentation, ViCLIP and VGG19 achieve the second-lowest performance, with InternVideo showing comparable but slightly lower scores. CLIP performs the worst among the models evaluated, which is likely due to its training being limited to static images, hindering its ability to capture the temporal dynamics necessary for video understanding. Notably, incorporating mask-augmented frames into our training pipeline leads to a significant improvement in the generalization ability of our model. Specifically, we observe a performance gain of approximately 5 percent points (pp.) on the Waymo dataset and 2 pp.\ on the NuScenes dataset, demonstrating the effectiveness of this augmentation strategy in enhancing temporal understanding.

\subsection{Generalization capabilities}
\label{sec: generalization}
\begin{figure}[t]
    \centering
    \includegraphics[width=0.4\textwidth]{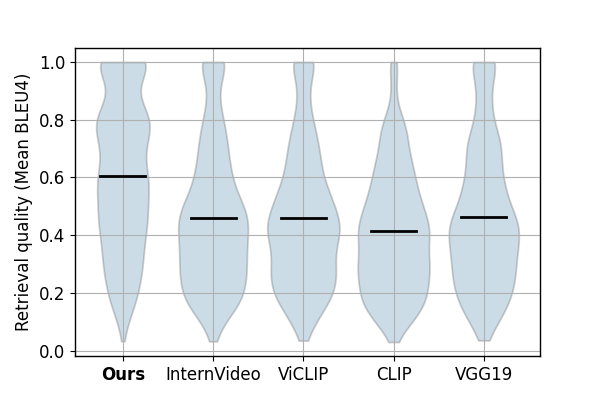}
    \caption{\textbf{Two-sentence captions retrieval.} The violin plots show the distribution of $\mathrm{BLEU4}$ scores between the ground truth captions of query and retrieved videos. A wider spread towards higher $\mathrm{BLEU4}$ scores indicates better retrieval quality. The thick black line shows the mean $\mathrm{BLEU4}$ of all the retrievals.}
    \label{fig: retrieval bleu distribution}
\end{figure}

We evaluate the generalization capability of our model by examining its zero-shot performance on more complex, two-sentence captions. Specifically, we assess retrieval quality using the video-to-video retrieval framework described in \cref{sec: model benchmarking}. \Cref{fig: retrieval bleu distribution} presents a violin plot showing the distribution of mean $\mathrm{BLEU4}$ scores between the ground truth captions of query videos and those of the retrieved candidates.
Although our model was trained exclusively on single-sentence captions---typically associated with simpler maneuvers---it significantly outperforms all baseline models in retrieving videos based on two-sentence captions. The violin plots illustrate the density of $\mathrm{BLEU4}$ scores for each method: a greater spread towards the top (values closer to 1) indicates stronger retrieval quality. Our model produces a broader and denser distribution in the higher $\mathrm{BLEU4}$ range, with a mean score approximately 20\% higher than the best-performing baseline. In contrast, the baseline models (InternVideo, ViCLIP, CLIP, and VGG19) exhibit more concentrated density in the lower half of the plot, indicating weaker alignment between the query and retrieved captions. These results highlight the robustness and transferability of our learned embeddings, even when evaluated on more and temporally complex descriptions.

\subsection{Embedding space analysis}
\label{sec: qualitative}

\begin{figure}[t]
    \centering
    \includegraphics[width=0.5\textwidth]{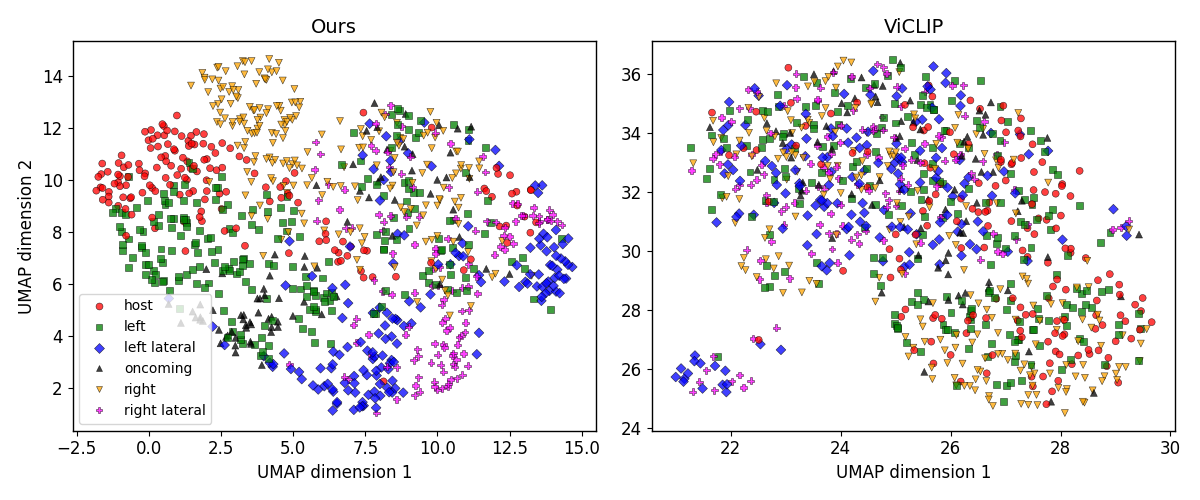}
    \caption{\textbf{UMAP representation of model's embeddings.} This figure illustrates the semantic structure of video embeddings produced by our model (left plot) and ViCLIP (right plot). Each point represents a video clip embedding, color-coded by its associated lane tag.
    %Our model's embeddings form more distinct and coherent clusters, reflecting stronger semantic organization and better alignment with lane-level spatial context.
    }
    \label{fig: umap retrieval neigbor}
\end{figure}

We investigate the semantic structure of the learned representations by visualizing the video embeddings using UMAP projections. \Cref{fig: umap retrieval neigbor} compares the embeddings generated by our model with those from ViCLIP. Each point corresponds to a video clip embedding, color-coded by its associated lane tag (e.g., host, left, right, left lateral, right lateral, oncoming).

The clusters corresponding to different lane tags are more clearly defined and spatially coherent in our model’s embedding space (left plot of \cref{fig: umap retrieval neigbor}). For instance, left, host and right lanes form distinct, well-localized regions in the embedding space, suggesting that the model has successfully learned to encode semantic differences tied to spatial roles in street scenes. This structured layout indicates that the embeddings capture meaningful representations aligned with lane semantics, contributing to improved interpretability and generalization performance, even though it was trained solely on single-sentence captions.
In contrast, the ViCLIP embeddings (right plot of \cref{fig: umap retrieval neigbor}) exhibit a much more entangled structure. The lane tag clusters are less well-formed and exhibit substantial overlap, indicating limited semantic disentanglement. This lack of structure suggests that ViCLIP embeddings are less aligned with the spatial semantics captured in the captions, which likely contributes to its weaker performance in retrieval and captioning tasks. Overall, these visualizations demonstrate the superiority of our model’s embedding space in preserving semantic structure, highlighting its stronger alignment between visual and linguistic representations.

\section{Conclusion}
We present a fully automated LiDAR-based captioning procedure that processes raw LiDAR sweeps into rich, temporally grounded captions. Since the method relies only on 3D point clouds, it can be plugged into any driving-scene dataset that contains LiDAR, instantly yielding large-scale, high-quality training data.

Using these auto-generated captions, we fine-tune the SwinBERT video captioning model and evaluate it on three datasets. To quantify temporal reasoning and exposure to “visual/static bias” (i.e., the tendency to rely on a single frame/background instead of motion cues), we introduce a custom metric, VBM. Lower VBM score indicates that the model attends more to dynamics such as velocity changes, lane shifts, or relative distance, rather than to a static features like background. Across all three datasets, VBM drops consistently, confirming that our captions enable the model to reason over time.

Finally, we push generalization further by supplying mask-augmented frames in which the LiDAR-derived object contour is overlaid directly on the frames. Since the masks are produced automatically by our convex-hull procedure, they scale with the data and require no human effort. The masked input forces SwinBERT to further focus on the object of interest rather than the background, enabling robust zero-shot captioning of unseen street scenes.

\bibliographystyle{IEEEtran}
\bibliography{references.bib}

% Generated by IEEEtran.bst, version: 1.14 (2015/08/26)
\begin{thebibliography}{10}
\providecommand{\url}[1]{#1}
\csname url@samestyle\endcsname
\providecommand{\newblock}{\relax}
\providecommand{\bibinfo}[2]{#2}
\providecommand{\BIBentrySTDinterwordspacing}{\spaceskip=0pt\relax}
\providecommand{\BIBentryALTinterwordstretchfactor}{4}
\providecommand{\BIBentryALTinterwordspacing}{\spaceskip=\fontdimen2\font plus
\BIBentryALTinterwordstretchfactor\fontdimen3\font minus \fontdimen4\font\relax}
\providecommand{\BIBforeignlanguage}[2]{{%
\expandafter\ifx\csname l@#1\endcsname\relax
\typeout{** WARNING: IEEEtran.bst: No hyphenation pattern has been}%
\typeout{** loaded for the language `#1'. Using the pattern for}%
\typeout{** the default language instead.}%
\else
\language=\csname l@#1\endcsname
\fi
#2}}
\providecommand{\BIBdecl}{\relax}
\BIBdecl

\bibitem{kuo2023mammutsimplearchitecturejoint}
\BIBentryALTinterwordspacing
W.~Kuo, A.~Piergiovanni, D.~Kim, X.~Luo, B.~Caine, W.~Li, A.~Ogale, L.~Zhou, A.~Dai, Z.~Chen, C.~Cui, and A.~Angelova, ``Mammut: A simple architecture for joint learning for multimodal tasks,'' 2023. [Online]. Available: \url{https://arxiv.org/abs/2303.16839}
\BIBentrySTDinterwordspacing

\bibitem{he2023vlabenhancingvideolanguage}
\BIBentryALTinterwordspacing
X.~He, S.~Chen, F.~Ma, Z.~Huang, X.~Jin, Z.~Liu, D.~Fu, Y.~Yang, J.~Liu, and J.~Feng, ``Vlab: Enhancing video language pre-training by feature adapting and blending,'' 2023. [Online]. Available: \url{https://arxiv.org/abs/2305.13167}
\BIBentrySTDinterwordspacing

\bibitem{chen2023valorvisionaudiolanguageomniperceptionpretraining}
\BIBentryALTinterwordspacing
S.~Chen, X.~He, L.~Guo, X.~Zhu, W.~Wang, J.~Tang, and J.~Liu, ``Valor: Vision-audio-language omni-perception pretraining model and dataset,'' 2023. [Online]. Available: \url{https://arxiv.org/abs/2304.08345}
\BIBentrySTDinterwordspacing

\bibitem{wang2019selfsupervisedspatiotemporalrepresentationlearning}
\BIBentryALTinterwordspacing
J.~Wang, J.~Jiao, L.~Bao, S.~He, Y.~Liu, and W.~Liu, ``Self-supervised spatio-temporal representation learning for videos by predicting motion and appearance statistics,'' 2019. [Online]. Available: \url{https://arxiv.org/abs/1904.03597}
\BIBentrySTDinterwordspacing

\bibitem{caron2019unsupervisedpretrainingimagefeatures}
\BIBentryALTinterwordspacing
M.~Caron, P.~Bojanowski, J.~Mairal, and A.~Joulin, ``Unsupervised pre-training of image features on non-curated data,'' 2019. [Online]. Available: \url{https://arxiv.org/abs/1905.01278}
\BIBentrySTDinterwordspacing

\bibitem{jaworek2019melanoma}
J.~Jaworek-Korjakowska, P.~Kleczek, and M.~Gorgon, ``Melanoma thickness prediction based on convolutional neural network with vgg-19 model transfer learning,'' in \emph{Proceedings of the IEEE/CVF Conference on Computer Vision and Pattern Recognition Workshops}, 2019, pp. 0--0.

\bibitem{xie2018pre}
Y.~Xie and D.~Richmond, ``Pre-training on grayscale imagenet improves medical image classification,'' in \emph{Proceedings of the European conference on computer vision (ECCV) workshops}, 2018, pp. 0--0.

\bibitem{jin2019low}
T.~Jin, S.~Huang, Y.~Li, and Z.~Zhang, ``Low-rank hoca: Efficient high-order cross-modal attention for video captioning,'' \emph{arXiv preprint arXiv:1911.00212}, 2019.

\bibitem{nakamura2021sensor}
K.~Nakamura, H.~Ohashi, and M.~Okada, ``Sensor-augmented egocentric-video captioning with dynamic modal attention,'' in \emph{Proceedings of the 29th ACM International Conference on Multimedia}, 2021, pp. 4220--4229.

\bibitem{ullah2021boosting}
N.~Ullah and P.~P. Mohanta, ``Boosting video captioning with dynamic loss network,'' \emph{arXiv preprint arXiv:2107.11707}, 2021.

\bibitem{kirillov2019panoptic}
A.~Kirillov, K.~He, R.~Girshick, C.~Rother, and P.~Doll{\'a}r, ``Panoptic segmentation,'' in \emph{Proceedings of the IEEE/CVF conference on computer vision and pattern recognition}, 2019, pp. 9404--9413.

\bibitem{divvala2009empirical}
S.~K. Divvala, D.~Hoiem, J.~H. Hays, A.~A. Efros, and M.~Hebert, ``An empirical study of context in object detection,'' in \emph{2009 IEEE Conference on computer vision and Pattern Recognition}.\hskip 1em plus 0.5em minus 0.4em\relax IEEE, 2009, pp. 1271--1278.

\bibitem{carion2020end}
N.~Carion, F.~Massa, G.~Synnaeve, N.~Usunier, A.~Kirillov, and S.~Zagoruyko, ``End-to-end object detection with transformers,'' in \emph{European conference on computer vision}.\hskip 1em plus 0.5em minus 0.4em\relax Springer, 2020, pp. 213--229.

\bibitem{wang2022internvideo}
Y.~Wang, K.~Li, Y.~Li, Y.~He, B.~Huang, Z.~Zhao, H.~Zhang, J.~Xu, Y.~Liu, Z.~Wang \emph{et~al.}, ``Internvideo: General video foundation models via generative and discriminative learning,'' \emph{arXiv preprint arXiv:2212.03191}, 2022.

\bibitem{wang2024internvideo2}
Y.~Wang, K.~Li, X.~Li, J.~Yu, Y.~He, G.~Chen, B.~Pei, R.~Zheng, J.~Xu, Z.~Wang \emph{et~al.}, ``Internvideo2: Scaling video foundation models for multimodal video understanding,'' \emph{arXiv preprint arXiv:2403.15377}, 2024.

\bibitem{cheng2024videollama}
Z.~Cheng, S.~Leng, H.~Zhang, Y.~Xin, X.~Li, G.~Chen, Y.~Zhu, W.~Zhang, Z.~Luo, D.~Zhao \emph{et~al.}, ``Videollama 2: Advancing spatial-temporal modeling and audio understanding in video-llms,'' \emph{arXiv preprint arXiv:2406.07476}, 2024.

\bibitem{lin2022swinbert}
K.~Lin, L.~Li, C.-C. Lin, F.~Ahmed, Z.~Gan, Z.~Liu, Y.~Lu, and L.~Wang, ``Swinbert: End-to-end transformers with sparse attention for video captioning,'' in \emph{Proceedings of the IEEE/CVF Conference on Computer Vision and Pattern Recognition}, 2022, pp. 17\,949--17\,958.

\bibitem{marculingoqa}
A.-M. Marcu, L.~Chen, J.~H{\"u}nermann, A.~Karnsund, B.~Hanotte, P.~Chidananda, S.~Nair, V.~Badrinarayanan, A.~Kendall, J.~Shotton \emph{et~al.}, ``Lingoqa: Visual question answering for autonomous driving.''

\bibitem{qian2024nuscenes}
T.~Qian, J.~Chen, L.~Zhuo, Y.~Jiao, and Y.-G. Jiang, ``Nuscenes-qa: A multi-modal visual question answering benchmark for autonomous driving scenario,'' in \emph{Proceedings of the AAAI Conference on Artificial Intelligence}, vol.~38, no.~5, 2024, pp. 4542--4550.

\bibitem{inoue2024nuscenes}
Y.~Inoue, Y.~Yada, K.~Tanahashi, and Y.~Yamaguchi, ``Nuscenes-mqa: Integrated evaluation of captions and qa for autonomous driving datasets using markup annotations,'' in \emph{Proceedings of the IEEE/CVF Winter Conference on Applications of Computer Vision}, 2024, pp. 930--938.

\bibitem{wu2023language}
D.~Wu, W.~Han, T.~Wang, Y.~Liu, X.~Zhang, and J.~Shen, ``Language prompt for autonomous driving,'' \emph{arXiv preprint arXiv:2309.04379}, 2023.

\bibitem{arai2024covla}
H.~Arai, K.~Miwa, K.~Sasaki, Y.~Yamaguchi, K.~Watanabe, S.~Aoki, and I.~Yamamoto, ``Covla: Comprehensive vision-language-action dataset for autonomous driving,'' \emph{arXiv preprint arXiv:2408.10845}, 2024.

\bibitem{caesar2020nuscenes}
H.~Caesar, V.~Bankiti, A.~H. Lang, S.~Vora, V.~E. Liong, Q.~Xu, A.~Krishnan, Y.~Pan, G.~Baldan, and O.~Beijbom, ``nuscenes: A multimodal dataset for autonomous driving,'' in \emph{Proceedings of the IEEE/CVF conference on computer vision and pattern recognition}, 2020, pp. 11\,621--11\,631.

\bibitem{sun2020scalability}
P.~Sun, H.~Kretzschmar, X.~Dotiwalla, A.~Chouard, V.~Patnaik, P.~Tsui, J.~Guo, Y.~Zhou, Y.~Chai, B.~Caine \emph{et~al.}, ``Scalability in perception for autonomous driving: Waymo open dataset,'' in \emph{Proceedings of the IEEE/CVF conference on computer vision and pattern recognition}, 2020, pp. 2446--2454.

\bibitem{liao2022kitti}
Y.~Liao, J.~Xie, and A.~Geiger, ``Kitti-360: A novel dataset and benchmarks for urban scene understanding in 2d and 3d,'' \emph{IEEE Transactions on Pattern Analysis and Machine Intelligence}, vol.~45, no.~3, pp. 3292--3310, 2022.

\bibitem{alexey2020image}
D.~Alexey, ``An image is worth 16x16 words: Transformers for image recognition at scale,'' \emph{arXiv preprint arXiv: 2010.11929}, 2020.

\bibitem{liu2021swin}
Z.~Liu, Y.~Lin, Y.~Cao, H.~Hu, Y.~Wei, Z.~Zhang, S.~Lin, and B.~Guo, ``Swin transformer: Hierarchical vision transformer using shifted windows,'' in \emph{Proceedings of the IEEE/CVF international conference on computer vision}, 2021, pp. 10\,012--10\,022.

\bibitem{touvron2021training}
H.~Touvron, M.~Cord, M.~Douze, F.~Massa, A.~Sablayrolles, and H.~J{\'e}gou, ``Training data-efficient image transformers \& distillation through attention,'' in \emph{International conference on machine learning}.\hskip 1em plus 0.5em minus 0.4em\relax PMLR, 2021, pp. 10\,347--10\,357.

\bibitem{bertasius2021space}
G.~Bertasius, H.~Wang, and L.~Torresani, ``Is space-time attention all you need for video understanding?'' in \emph{ICML}, vol.~2, no.~3, 2021, p.~4.

\bibitem{arnab2021vivit}
A.~Arnab, M.~Dehghani, G.~Heigold, C.~Sun, M.~Lu{\v{c}}i{\'c}, and C.~Schmid, ``Vivit: A video vision transformer,'' in \emph{Proceedings of the IEEE/CVF international conference on computer vision}, 2021, pp. 6836--6846.

\bibitem{zhao2022tuber}
J.~Zhao, Y.~Zhang, X.~Li, H.~Chen, B.~Shuai, M.~Xu, C.~Liu, K.~Kundu, Y.~Xiong, D.~Modolo \emph{et~al.}, ``Tuber: Tubelet transformer for video action detection,'' in \emph{Proceedings of the IEEE/CVF Conference on Computer Vision and Pattern Recognition}, 2022, pp. 13\,598--13\,607.

\bibitem{wang2021end}
Y.~Wang, Z.~Xu, X.~Wang, C.~Shen, B.~Cheng, H.~Shen, and H.~Xia, ``End-to-end video instance segmentation with transformers,'' in \emph{Proceedings of the IEEE/CVF conference on computer vision and pattern recognition}, 2021, pp. 8741--8750.

\bibitem{zhang2023video}
H.~Zhang, X.~Li, and L.~Bing, ``Video-llama: An instruction-tuned audio-visual language model for video understanding,'' \emph{arXiv preprint arXiv:2306.02858}, 2023.

\bibitem{lin2023video}
B.~Lin, Y.~Ye, B.~Zhu, J.~Cui, M.~Ning, P.~Jin, and L.~Yuan, ``Video-llava: Learning united visual representation by alignment before projection,'' \emph{arXiv preprint arXiv:2311.10122}, 2023.

\bibitem{jin2023adapt}
B.~Jin, X.~Liu, Y.~Zheng, P.~Li, H.~Zhao, T.~Zhang, Y.~Zheng, G.~Zhou, and J.~Liu, ``Adapt: Action-aware driving caption transformer,'' in \emph{2023 IEEE International Conference on Robotics and Automation (ICRA)}.\hskip 1em plus 0.5em minus 0.4em\relax IEEE, 2023, pp. 7554--7561.

\bibitem{kim2018textual}
J.~Kim, A.~Rohrbach, T.~Darrell, J.~Canny, and Z.~Akata, ``Textual explanations for self-driving vehicles,'' in \emph{Proceedings of the European conference on computer vision (ECCV)}, 2018, pp. 563--578.

\bibitem{wang2023internvid}
Y.~Wang, Y.~He, Y.~Li, K.~Li, J.~Yu, X.~Ma, X.~Li, G.~Chen, X.~Chen, Y.~Wang \emph{et~al.}, ``Internvid: A large-scale video-text dataset for multimodal understanding and generation,'' \emph{arXiv preprint arXiv:2307.06942}, 2023.

\bibitem{radford2021learning}
A.~Radford, J.~W. Kim, C.~Hallacy, A.~Ramesh, G.~Goh, S.~Agarwal, G.~Sastry, A.~Askell, P.~Mishkin, J.~Clark \emph{et~al.}, ``Learning transferable visual models from natural language supervision,'' in \emph{International conference on machine learning}.\hskip 1em plus 0.5em minus 0.4em\relax PMLR, 2021, pp. 8748--8763.

\bibitem{simonyan2014very}
K.~Simonyan and A.~Zisserman, ``Very deep convolutional networks for large-scale image recognition,'' \emph{arXiv preprint arXiv:1409.1556}, 2014.

\bibitem{fukuzawa2025can}
T.~Fukuzawa, K.~Hara, H.~Kataoka, and T.~Tamaki, ``Can masking background and object reduce static bias for zero-shot action recognition?'' in \emph{International Conference on Multimedia Modeling}.\hskip 1em plus 0.5em minus 0.4em\relax Springer, 2025, pp. 366--379.

\bibitem{andrew1979another}
A.~M. Andrew, ``Another efficient algorithm for convex hulls in two dimensions,'' \emph{Information Processing Letters}, vol.~9, no.~5, pp. 216--219, 1979.

\bibitem{papineni2002bleu}
K.~Papineni, S.~Roukos, T.~Ward, and W.-J. Zhu, ``Bleu: a method for automatic evaluation of machine translation,'' in \emph{Proceedings of the 40th annual meeting of the Association for Computational Linguistics}, 2002, pp. 311--318.

\end{thebibliography}

\section{Supplementary Material}
In the next section, we detail the implementation of the various thresholds and the corresponding data used to extract the necessary information for constructing the final neighbor captions. \Cref{sec: host tagging} outlines the implementation of host motion and direction tagging, which supports host caption construction. Meanwhile, \Cref{sec: stationary tagging}, \Cref{sec: lane tagging}, \Cref{sec: motion tagging}, and \Cref{sec: unified tagging} describe how the relevant information is generated for neighbor caption construction.

\subsection{Host tagging}
\label{sec: host tagging}
This is the first stage of the rule-based captioning system. Initially, the host’s velocity and yaw rate are extracted from the sensor data. At each timestamp, thresholds (see \cref{eq: host motion label} and \cref{eq: host direction label}) are applied to assign the corresponding motion and direction tags.

\begin{equation}
    \label{eq: host motion label}
    \text{Motion tag} = 
    \begin{cases}
    \text{Stationary}, & \text{if } v < T_v \\
    \text{Accelerating}, & \text{if } \frac{dv}{dt} > T_a \\
    \text{Decelerating}, & \text{if } \frac{dv}{dt} < -T_a \\
    \text{Cruising}, & \text{if } -T_a <= \frac{dv}{dt} <= T_a
    \end{cases}
\end{equation}
where, $v$, $t$ and $\frac{dv}{dt}$ denote the host vehicle's instantaneous velocity, time between two successive velocity recordings, and the host's instantaneous acceleration, respectively.  $T_v$ and $T_a$ represent the thresholds for the velocity and acceleration of the host vehicle.

\begin{equation}
    \label{eq: host direction label}
    \text{Direction tag} = 
    \begin{cases}
    \text{Steering right}, & \text{if } w < -T_w \\
    \text{Steering left}, & \text{if } w > T_w \\
    \text{Heading straight}, & \text{if } -T_w <= w <= T_w
    \end{cases}
\end{equation}
where, $w$ and $t$ represent the yaw rate and the time between two successive yaw rate measurements of the host vehicle. $T_w$ is the threshold on the yaw rate of the host vehicle.

\subsection{Non-Stationary neighbor selection}
\label{sec: stationary tagging}

As explained in the methodology, a key step in caption generation involves extracting object class, lane tags, and motion tags from object tracks, obtained via a SOTA 3D LiDAR–based detector and tracker, on raw LiDAR data. These tags are generated exclusively for non-stationary neighbors, as determined by \cref{eq: stationary}. A neighbor is deemed stationary by thresholding changes in its bounding box position relative to the host’s initial frame, referred to as the host-compensated location. The transformation matrix for computing this host-compensated location is derived from the LiDAR sensor’s calibration data and the host vehicle’s position relative to its starting location.
\begin{equation}
    \label{eq: stationary}
    Stationary = 
    \begin{cases}
    \text{True}, & \text{if } \frac{d{h_x}}{dt} < T_s \text{and} \frac{d{h_y}}{dt} < T_s \\
    \text{False}, & \text{otherwise}
    \end{cases}
\end{equation}

here, $h_x$ and $h_y$ denote the object’s center distances from the host vehicle’s position at the first frame, measured along and perpendicular to the host’s heading at that frame, respectively. The threshold $T_s$ determines whether a neighbor is classified as stationary. Once non-stationary neighbors are identified, the tagging procedures are carried out as described in \cref{sec: lane tagging}, \cref{sec: motion tagging}, and \cref{sec: unified tagging}.

\subsection{Lane tagging}
\label{sec: lane tagging}

In this section, we detail the implementation of the lane tagging algorithm used for caption construction. The algorithm is divided into two parts: the baseline lane tag and the lane tag itself. The baseline step identifies neighbors in oncoming, lateral (lanes perpendicular to the host lane, such as at intersections), or ongoing lanes by thresholding each neighbor’s yaw, as shown in \cref{eq: lane tag baseline}. The second step locates each neighbor relative to the host, classifying them as left, right, or host lane by combining the baseline lane tag with the neighbor’s bounding box center in the y-direction, as specified in \cref{eq: lane tag}.

\begin{equation}
    \label{eq: lane tag baseline}
    L_B = 
    \begin{cases}
    \text{oncoming}, & \text{if } \phi_{o_l} < \phi < \phi_{o_u} \\
    \text{lateral}, & \text{if } \phi_{l_l} < \phi < \phi_{l_u} \\
    \text{ongoing}, & \text{otherwise}
    \end{cases}
\end{equation}

where $\phi$ denotes the neighbor's yaw, and $L_B$ represents the baseline lane tag, which is refined in a subsequent step to obtain the final lane tags. The parameters $\phi_{o_l}$ and $\phi_{o_u}$ are the lower and upper yaw thresholds for identifying neighbors on the oncoming lane, while $\phi_{l_l}$ and $\phi_{l_u}$ specify the yaw thresholds for identifying neighbors on the lateral lane.  

\begin{equation}
    \label{eq: lane tag}
    L =   
    \begin{cases}
    \text{right lateral}, & \text{if } L_B = \text{lateral and } y > T_h \\
    \text{left lateral}, & \text{if } L_B = \text{lateral and } y < -T_h \\
    \text{host lateral}, & \text{if } L_B = \text{lateral and otherwise}  \\
    \text{right}, & \text{if } L_B = \text{ongoing and } y > T_h \\
    \text{left}, & \text{if } L_B = \text{ongoing and } y < -T_h \\
    \text{host}, & \text{if } L_B = \text{ongoing and otherwise}  \\
    \text{oncoming}, & \text{otherwise}
    \end{cases}
\end{equation}
where, $L$ and $L_B$ are the refined lane tag and the baseline lane tag respectively. $y$ denotes the distance of object's center from host vehicle perpendicular to the host's heading direction and $T_h$ is the threshold for determining whether a neighbor is on the host lane.

\subsection{Motion tagging}
Another key step is to classify the neighbor’s motion relative to the host, using the previously determined lane tags along with a threshold on the neighbor’s lateral position relative to the host’s heading. 
\label{sec: motion tagging}
\begin{equation}
    \label{eq: motion tag}
    M = 
    \begin{cases}
    \text{approach}, & \text{if } L = \text{left lateral and } \frac{dy}{dt} > T_y \\
    \text{away}, & \text{if } L = \text{left lateral and } |\frac{dy}{dt}| > T_y \\
    \text{constant}, & \text{if } L = \text{left lateral and otherwise} \\
    \text{away}, & \text{if } L = \text{right lateral and } \frac{dy}{dt} > T_y \\
    \text{approach}, & \text{if } L = \text{right lateral and } |\frac{dy}{dt}| > T_y \\
    \text{constant}, & \text{if } L = \text{right lateral and otherwise} \\
    \text{stationary}, & \text{if } L = \text{lateral and otherwise}  \\
    \text{constant}, & \text{otherwise}
    \end{cases}
\end{equation}
where $M$ and $L$ denote the motion and lane tags, respectively. The term $\frac{dy}{dt}$ represents the neighbor's instantaneous velocity in the direction perpendicular to the host's heading and $T_y$ is the threshold used to determine whether a neighbor maintains a constant distance from the host. 

\subsection{Unified tag generation}
\label{sec: unified tagging}

After extracting the lane and motion tags, we concatenate them across time to form a timeseries of concatenated tags. This time-series is then “unified” into a unique sequence of concatenated tags (retaining their order of occurrence), referred to as the unified tag sequence, which captures the neighbor’s action within that time window.
\begin{equation}
\begin{aligned}
    \label{eq: unified tag timeseries}
    &c_i = {l_i}{m_i} \mid l_i \in L, m_i \in M, i \in [0, n] \\
    &C = [c_0, c_1, c_2, \dots, c_n] \\
    &U = [c_i, \text{if } c_i \neq c_{i+1}, i \in [0, n]]
\end{aligned}
\end{equation}
where $c_i$, $l_i$ and $m_i$ are the concatenated, lane and motion tags at timestep $i$ respectively. $L$,$M$, $C$ and $U$ are the sequence of the neighbor's lane, motion, concatenated and unified tags respectively.

\subsection{Extended study on embeddings}
\label{sec: embedding extended}

\begin{figure}[t]
    \centering
    \includegraphics[width=0.49\textwidth]{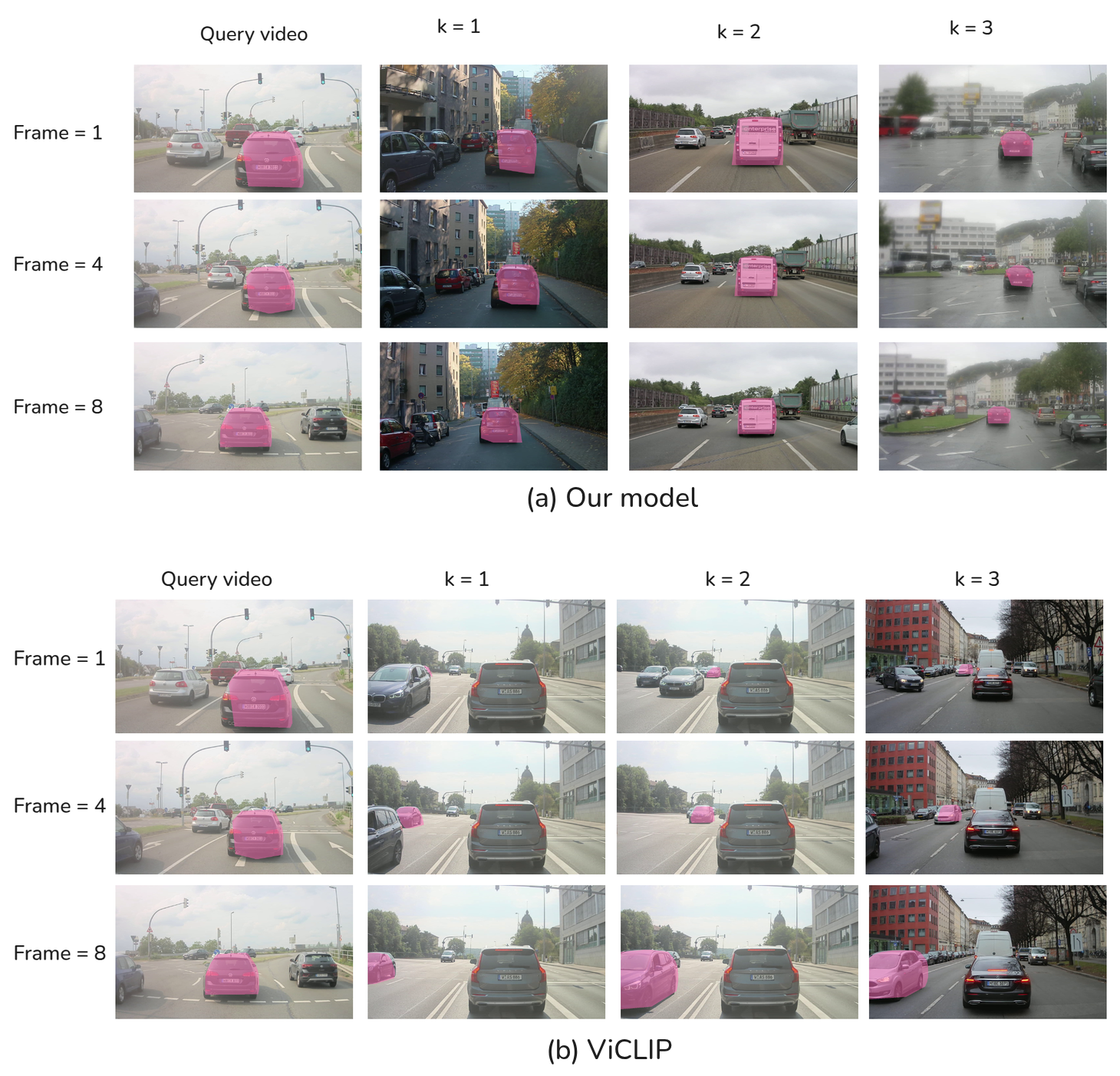}
    \caption{\textbf{Video-to-video retrieval.} This figure shows an example of retrieving the top-three nearest neighbors for a given query video. The first column shows the query frame, and the second, third, and fourth columns present the first, second, and third nearest neighbor frames, respectively. The pink mask highlights the object of interest, which is the focus of each frame’s caption. Figure 1a and Figure 1b depict the retrieval results using our model’s embeddings and ViCLIP’s embeddings, respectively.}
    \label{fig: retrieval neigbor}
\end{figure}

To further examine our model’s understanding of temporal dynamics, we qualitatively evaluate its performance in a video-to-video retrieval task and compare it with that of ViCLIP. \Cref{fig: retrieval neigbor} presents the top three nearest neighbor videos retrieved for a given query, using embeddings from both our model and ViCLIP. In the figure, the first column shows frames from the query video, while the second, third, and fourth columns display frames from the top three retrieved neighbors.

The results reveal a notable difference between the two models. Our model retrieves videos where the object of interest is consistently performing the same action as in the query, even though the surrounding visual context such as lighting conditions, background scenery, and weather varies significantly. This suggests that our model focuses more on the temporal and semantic content of the video, rather than being influenced by low-level visual similarities. In contrast, ViCLIP’s retrieved neighbors often appear more visually similar in appearance but are less aligned in terms of the underlying action. These observations indicate that our model exhibits reduced sensitivity to visual biases and demonstrates a stronger temporal understanding.

\end{document}